\documentclass[10pt, a4paper]{article}
\usepackage{lrec2022} 
\usepackage{multibib}
\newcites{languageresource}{Language Resources}
\usepackage{graphicx}

\usepackage{subcaption}

\usepackage{tabularx}
\usepackage{soul}
\usepackage{titlesec}
\titleformat{\section}{\normalfont\large\bfseries\center}{\thesection.}{1em}{}
\titleformat{\subsection}{\normalfont\SmallTitleFont\bfseries\raggedright}{\thesubsection.}{1em}{}
\titleformat{\subsubsection}{\normalfont\normalsize\bfseries\raggedright}{\thesubsubsection.}{1em}{}
\renewcommand\thesection{\arabic{section}}
\renewcommand\thesubsection{\thesection.\arabic{subsection}}
\renewcommand\thesubsubsection{\thesubsection.\arabic{subsubsection}}

\usepackage{epstopdf}
\usepackage[utf8]{inputenc}

\usepackage{hyperref}
\usepackage{xstring}

\usepackage{color}

\usepackage{adjustbox}
\usepackage{lipsum}
\usepackage{tikz}
\newcommand*\circled[1]{\tikz[baseline=(char.base)]{
            \node[shape=circle,draw,inner sep=2pt] (char) {#1};}}

\RequirePackage{color}

\title{Evaluation Benchmarks for Spanish Sentence Representations}

 \name{Vladimir Araujo\textsuperscript{1,2,6,7},
 Andr\'es Carvallo\textsuperscript{1,6},
 Souvik Kundu\textsuperscript{3},
 Jos\'e Cañete\textsuperscript{4,6}, \\
 {\bf \large
 Marcelo Mendoza\textsuperscript{5,6,7},
 Robert E. Mercer\textsuperscript{3},
 Felipe Bravo-Marquez\textsuperscript{4,6,7},
 }\\
 {\bf \large
 Marie-Francine Moens\textsuperscript{2}, 
 Alvaro Soto\textsuperscript{1,6}
 }}

 \address{\textsuperscript{1}Pontificia Universidad Cat\'olica de Chile, Santiago, Chile \\
\textsuperscript{2}KU Leuven, Leuven, Belgium \\
 \textsuperscript{3}University of Western Ontario, London, Canada \\
 \textsuperscript{4}Department of Computer Science, University of Chile, Santiago, Chile \\
 \textsuperscript{5}Universidad Técnica Federico Santa María, Santiago, Chile \\
 \textsuperscript{6} National Center for Artificial Intelligence Research (CENIA), Santiago, Chile \\
\textsuperscript{7} Millennium Institute for Foundational Research on Data (IMFD), Santiago, Chile \\
 \href{mailto:vgaraujo@uc.cl}{\texttt{vgaraujo@uc.cl}}\\}

\abstract{
Due to the success of pre-trained language models, versions of languages other than English have been released in recent years.
This fact implies the need for resources to evaluate these models.
In the case of Spanish, there are few ways to systematically assess the models' quality.
In this paper, we narrow the gap by building two evaluation benchmarks.
Inspired by previous work \cite{conneau-kiela-2018-senteval,chen-etal-2019-evaluation}, we introduce Spanish SentEval and Spanish DiscoEval, aiming to assess the capabilities of stand-alone and discourse-aware sentence representations, respectively. 
Our benchmarks include considerable pre-existing and newly constructed datasets that address different tasks from various domains.
In addition, we evaluate and analyze the most recent pre-trained Spanish language models to exhibit their capabilities and limitations. As an example, we discover that for the case of discourse evaluation tasks, mBERT, a language model trained on multiple languages, usually provides a richer latent representation than models trained only with documents in Spanish. 
We hope our contribution will motivate a fairer, more comparable, and less cumbersome way to evaluate future Spanish language models.
 \\ \newline  }

\begin{document}

\maketitleabstract

\section{Introduction}

Spanish is one of the most widely spoken languages.
This fact has drawn the attention of the NLP community to the development of resources for that language.
As a result, some pre-trained Spanish language models \cite{etcheverry-wonsever-2016-spanish,che-etal-2018-towards,canete2020spanish,gut2021spanish} have been released in recent years driven by self-supervised approaches.
This proliferation of Spanish language models increases the need for annotated datasets to evaluate them. Consequently, Spanish datasets for a wide variety of independent tasks have been proposed \cite{a-garcia-cumbreras-etal-2006-bruja,cruz2008clasificacion,artetxe-etal-2020-translation,huertastato2021silt}.
However, little effort has been put into creating benchmarks that allow models to be evaluated systematically and fairly.

Recently, \newcite{canete2020spanish} presented the GLUES benchmark, a compilation of natural language understanding tasks in Spanish. 
This benchmark aims to evaluate the performance of models by fine-tuning them to a target task \cite{wang-etal-2018-glue}.
In contrast, another methodology known as probing tasks aims to assess whether the resulting representations of the models are general-purpose \cite{conneau-etal-2018-cram}. 
A probing task is designed in such a way as to isolate some linguistic phenomena, and a classifier is used on top of the representations to verify if the model has encoded the linguistic phenomena in question. 
This type of representation evaluation for Spanish language models is generally carried out using a cross-lingual setting \cite{ravishankar-etal-2019-probing,10.1162/coli_a_00376}. However, these benchmarks only focus on assessing word representations or basic linguistic knowledge.

On the one hand, the \textbf{Spanish SentEval}, inspired by SentEval \cite{conneau-kiela-2018-senteval}, aims to evaluate representations of independent sentences.
Unlike previous work focused on probing tasks for basic linguistic properties \cite{ravishankar-etal-2019-probing,10.1162/coli_a_00376}, our benchmark comprises four sets of sentence classification tasks with realistic texts from different domains.
On the other hand, the \textbf{Spanish DiscoEval}, inspired by DiscoEval \cite{chen-etal-2019-evaluation}, focuses on the evaluation of discourse knowledge in sentence representations. 
Evaluating discourse involves analyzing a sentence in the context in which it is located. For this reason, we include five sets of tasks based on sentence ordering, discourse relations, and discourse coherence.

The overall objective of both benchmarks is to avoid unnecessary re-implementations and the use of multiple evaluation schemes, thus allowing a comparable and fair assessment between models.
Furthermore, we compare publicly available Spanish sentence encoders on our Spanish SentEval and Spanish DiscoEval to demonstrate their strengths and weaknesses. The results and subsequent analysis expose the Spanish language models' current capabilities, showing that there is still room to improve them in future work.
Our code and datasets are available for future experimentation and replicability at \url{https://github.com/OpenCENIA/Spanish-Sentence-Evaluation}.

\section{Sentence Evaluation}
SentEval was introduced by \newcite{conneau-kiela-2018-senteval} as a tool for evaluating the quality of universal sentence representations. 
It encompasses a standard pipeline evaluation that uses the representations generated by sentence encoders as features in various downstream tasks. 
Specifically, SentEval includes stand-alone sentence and sentence pair tasks modeled by classification or regression. For comparison purposes, this framework consists of simple predefined neural architectures to avoid shifting the burden of modeling to their optimization process. 

For our Spanish SentEval, we adopt the original pipeline and include datasets equivalent to those in English.
Below we describe each task and dataset included in our Spanish version. Additionally, basic statistics for each dataset are shown in \hyperref[appendix:a]{Appendix A}.

As proposed in \newcite{chen-etal-2019-evaluation}, we use $[\cdot, \cdot, \cdots]$ to denote concatenation of vectors, $\odot$ for element-wise multiplication, and $| \cdot |$ for element-wise absolute value.

\subsection{Sentence Classification (SC)}
Sentence classification is one of the most common NLP tasks, with applications ranging from document classification to sentiment analysis. Because of its inherent simplicity, the task offers a straightforward way to evaluate sentence-level representations.
For our version, we include a set of binary and multiclass datasets that cover various types of sentence classification tasks.

For sentiment analysis, we include MuchoChine (MC) \cite{cruz2008clasificacion}, a movie review dataset, and TASS 2020 \cite{VegaDCAMZCACCM20} tasks 1 and 2 consisting of polarity and emotion classification \cite{plaza-del-arco-etal-2020-emoevent}, respectively.
Figure~\ref{fig.sc} shows an example of a MC positive sentiment sentence.

Other types of text classification datasets that we include are the FilmAffinity corpus (FAC) \cite{7360018} for subjective/objective classification and the Spanish QC dataset (SQC) \cite{a-garcia-cumbreras-etal-2006-bruja} for question-type. 

For all of these tasks, the input to the classifier is the representation of the sentence.

\begin{figure}[!h]
\begin{center}
\fbox
{
\begin{minipage}{0.45\textwidth}
• Una historia policiaca que Scorsese la transforma en una memorable muestra del genero.
\end{minipage}
}
\vspace{-0.2cm}
\caption{SC example. The sentence belongs to the MC dataset and shows a \textit{positive} sentiment.}
\vspace{-0.5cm}
\label{fig.sc}
\end{center}
\end{figure}

\subsection{Sentence Pair Classification (SPC)}
In sentence pair classification, each example in a dataset has two sentences along with the appropriate target, and the aim is to model the textual interaction between them.
We consider entailment and paraphrasing tasks for our Spanish benchmark.

For the entailment task, we include two datasets.
The first is the recently released SICK-es \cite{huertastato2021silt} for entailment (SICK-es-E), which was constructed by translating and manually curating the English SICK dataset into Spanish. 
Due to the lack of NLI tasks in Spanish, the second dataset was constructed using XNLI \cite{conneau-etal-2018-xnli} and esXNLI \cite{artetxe-etal-2020-translation}. Specifically, we use the XNLI test set as the training set, the XNLI development set as the development set, and the esXNLI set as the test set. We will refer to this as NLI-es (example shown in Figure~\ref{fig.spc}).

For the paraphrasing task, we use PAWS-X \cite{yang-etal-2019-paws}, a cross-lingual paraphrase identification dataset with high lexical overlap. We only use the Spanish text, naming it PAWS-es for ease of reference.

Like English SentEval, we encode the two sentences and use $[|x_{1} - x_{2}|, x_{1} \odot x_{2}]$ as input to the classifier.

\begin{figure}[!h]
\begin{center}
\fbox
{
\begin{minipage}{0.45\textwidth}
    \textit{\textbf{Premise}}: Y yo estaba bien, ¡y eso fue todo! \newline
    \textit{\textbf{Hypothesis}}: Después de que dije que sí, terminó.
\end{minipage}
}
\vspace{-0.2cm}
\caption{Example of SPC from NLI-es. The two sentences show an \textit{entailment}.}
\vspace{-0.5cm}
\label{fig.spc}
\end{center}
\end{figure}

\subsection{Semantic Similarity (SS)}
This task consists of scoring a pair of sentences based on their degree of similarity, even if they are not exact matches.
There are two common approaches to evaluating this task and we include them in our Spanish SentEval.
The first requires training a model on top of the sentence embeddings. For this approach, we use the SICK-es \cite{huertastato2021silt} for relatedness (SICK-es-R).
The second assesses sentence pairs using an unsupervised approach. For this case, we include the Spanish track of STS tasks 2014 \cite{agirre-etal-2014-semeval}, 2015 \cite{agirre-etal-2015-semeval} and 2017 \cite{cer-etal-2017-semeval}. 
All of these datasets consist of a pair of sentences labeled with a similarity score between 0 and 5; an example is shown in Figure~\ref{fig.ss}.
The objective is to evaluate whether the cosine similarity of two sentence representations correlates with a human-labeled similarity score through Pearson and Spearman correlations.

\begin{figure}[!h]
\begin{center}
\fbox
{
\begin{minipage}{0.45\textwidth}
    • Un perro está con un juguete. \newline
    • Un perro tiene un juguete.
\end{minipage}
}
\vspace{-0.2cm}
\caption{Example of the STS task. The two sentences are similar with a \textit{score} of 4.8 out of 5.}
\vspace{-0.5cm}
\label{fig.ss}
\end{center}
\end{figure}

\subsection{Linguistic Probing Tasks (LPT)}
Some sentence classification tasks are complex and make it difficult to infer what kind of information is present in the representations.
This prompted the creation of X-Probe \cite{ravishankar-etal-2019-probing}, a multilingual benchmark of nine probing tasks to evaluate individual linguistic properties.
These tasks were designed to evaluate surface information (SentLen, WC), syntactic information (BiShift, TreeDepth), and semantic information (Tense, SubjNum, ObjNum, SOMO, CoordInv).
The former evaluate superficial tasks that could be solved simply by looking at the sentence tokens. The second tests whether the embeddings are sensitive to the syntactic properties of the sentences. The third assesses the semantic understanding of the embedding.
We include all the proposed probing tasks in Spanish from X-Probe. We refer to the original paper \cite{ravishankar-etal-2019-probing} for further information. 

The input to the classifier is the representation of the sentence, and the output can be binary or multiclass. 

\begin{figure}[!h]
\begin{center}
\fbox
{
\begin{minipage}{0.45\textwidth}
• En enero participó en la infructuosa defensa de Forlí frente a César Borgia.
\end{minipage}
}
\vspace{-0.2cm}
\caption{Example of LPT. The task consists of Tense classification. In this case the sentence is in \textit{past tense}.}
\vspace{-0.5cm}
\label{fig.pt}
\end{center}
\end{figure}

\section{Discourse Evaluation}

DiscoEval originally proposed by \newcite{chen-etal-2019-evaluation} includes tasks to evaluate discourse-related knowledge in pretrained sentence representations. 
DiscoEval adopts the SentEval pipeline with fixed standard hyperparameters to avoid discrepancies.
For our Spanish version of DiscoEval, we follow closely the original construction and evaluation methodology. Specifically, DiscoEval includes supervised sentence and sentence group classification tasks modeled by logistic regression or classification.
Our datasets were constructed from multiple domains encompassing a wide diversity of text sources.
Below we describe the tasks and dataset constructions.
Statistics for each dataset are shown in \hyperref[appendix:a]{Appendix A}.

\subsection{Sentence Position (SP)}
SP seeks to assess the ability of a model to order ideas in a paragraph. 
This dataset is constructed by taking five consecutive sentences from a given corpus and randomly moving one of these five sentences to the first position.
The task consists of predicting the proper location of the first sentence. We have five classes where class 1 means that the first sentence is in the correct position. But if the class is between 2 and 5, the first sentence corresponds to another position in the paragraph.
We create three Spanish versions of different domains for this task using: 
the first five sentences of Wikipedia articles\footnote{We use the latest Spanish Wikipedia articles dump (\href{https://dumps.wikimedia.org/eswiki/latest/}{dumps.wikimedia.org/eswiki/latest/})}, Chilean university thesis abstracts\footnote{We collected abstracts from public repositories of the Pontificia Universidad Católica de Chile (\href{https://repositorio.uc.cl/}{repositorio.uc.cl}) and Universidad de Chile (\href{https://repositorio.uchile.cl/}{repositorio.uchile.cl}).}, and news articles in Spanish from the MLSUM dataset \cite{scialom-etal-2020-mlsum}.

Figure \ref{fig.sp} shows an example of this task for the thesis dataset. The first sentence should be in the second position among these sentences. To make correct predictions, the model needs to be aware of both typical orderings of events and how events are described in language. In the example shown, the model needs to understand that the objective of the thesis has to be described before the main findings of the study.

As proposed by \newcite{chen-etal-2019-evaluation} to train the classifier for this task, we first encode the five sentences into vector representations $x_i$. As input to the classifier, we concatenate $x_1$ and $x_1 - x_i$ for $2 \leq i \leq 5$: $[x_1, x_1 - x_2, x_1 -x_3 , x_1 - x_4 , x_1 - x_5]$. The output is between $1$ and $5$, which indicates the target position of the first sentence.

\begin{figure}[!h]
\begin{center}
\fbox
{
\begin{minipage}{0.45\textwidth}
\small
    1) Se encontró que la adición de nanopartículas de sílice aumenta la rigidez del material. \circled{2} \newline
    2) El objetivo de este trabajo es estudiar el efecto de la incorporación de nanopartículas de sílice en la rigidez de material. \newline
    3) Las Nanopartículas de sílice fueron sintetizadas utilizando el método sol-gel. \newline
    4) Las Nanopartículas de menor tamaño tienen un mayor efecto sobre las propiedades del material. \newline
    5) La rigidez del material aumentó hasta en un 80\% con la adición de 30\% de nanopartículas de silice.
\end{minipage}
}
\vspace{-0.2cm}
\caption{SP example of thesis domain. The number inside the circle shows the correct position of the first sentence. This sentence belongs in the 2nd place.}
\vspace{-0.5cm}
\label{fig.sp}
\end{center}
\end{figure}

\subsection{Binary Sentence Ordering (BSO)}
BSO is a binary classification task to determine if the order of two sentences is correct.
This task aims to assess the ability of sentence representations to capture local discourse coherence.
This data comes from the same three domains of the SP task. However, in this case, we only take the first two sentences of each text.

Figure \ref{fig.bso} provides an example from the Spanish Wikipedia. The order of the sentences is incorrect as the ``Neue Pinakothek'' museum should be mentioned before describing the art found inside.
In order to find the incorrect ordering in this example, the sentence representations need to be able to provide information if one sentence comes after or before the separation. 

As English DiscoEval to create the model inputs to train the classifiers, we concatenate the embeddings generated by the sentence encoder of both sentences with their element-wise difference: $[x_1, x_2, x_1-x_2]$.

\begin{figure}[!h]
\begin{center}
\fbox
{
\begin{minipage}{0.45\textwidth}
1) Se centra en el Arte europeo del siglo XIX. \newline 
2) El Neue Pinakothek es un museo de arte situado en Múnich, Alemania.
\end{minipage}
}
\vspace{-0.2cm}
\caption{Example from the Wikipedia domain of the BSO task. The sequence is in the wrong order.}
\vspace{-0.5cm}
\label{fig.bso}
\end{center}
\end{figure}

\subsection{Discourse Coherence (DC)}

The Discourse Coherence (DC) task is a sentence disentanglement task proposed to determine if a sequence of six sentences forms a coherent paragraph. We create three versions of this task, two from open-domain dialogue datasets and the other from Wikipedia articles. Given six coherent contiguous sentences, we randomly replace one of them with a sentence from another sequence. Note that we choose the sentence to replace uniformly among positions 2-5. We generate balanced datasets with coherent (positive) and non-coherent (negative) instances, which results in a binary classification task.

For the open-domain dialogue dataset, we use the OpenSubtitles\footnote{\href{http://www.opensubtitles.org/}{http://www.opensubtitles.org/}} corpus \cite{lison-tiedemann-2016-opensubtitles2016} and the Gutenberg Dialogue dataset \cite{csaky-recski-2021-gutenberg}. OpenSubtitles is a large corpus, so we randomly retrieve some dialogues and create the splits. In the case of the Gutenberg Dialogue, we use the original splits provided by the author. For the Wikipedia domain, we take only one coherent text from each article. Then we randomly create the splits. In all cases, we discard paragraphs with fewer than six sentences and we select the negative sample from other dialogues or articles in the corresponding domain. Figure~\ref{fig.dc} shows a dialog to which the second sentence does not belong.

Like English DiscoEval, we encode the six sentences as vector representations and concatenate them ($[x_1, x_2, x_3, x_4, x_5, x_6]$) as input to the classifier. 

\begin{figure}[!h]
\begin{center}
\fbox
{
\begin{minipage}{0.45\textwidth}
    1) ¡nicolás, ha llegado tu hora! \newline
    2) \textbf{recuerdo que en la galería obscura me  ofrecísteis vuestra casa.} \newline
    3) no; prefiero fumarme una pipa. \newline
    4) ¿dónde está tu pipa? \newline
    5) en el chaleco. \newline
    6) bien; aquí la tienes.
\end{minipage}
}
\vspace{-0.2cm}
\caption{Example of DC from the Gutenberg. The sentence in bold does not belong to the dialogue.}
\vspace{-0.5cm}
\label{fig.dc}
\end{center}
\end{figure}

\subsection{Sentence Section Prediction (SSP)}
SSP is a task to determine the section of a given sentence. This is based on the fact that the writing style can vary throughout a document, showing distinct patterns. 
The English DiscoEval originally used abstract and other sections of scientific papers to build the dataset. For our Spanish version, we use news articles instead. The news usually has a headline that is a sentence that presents the main idea of the article, a subhead that is a group of sentences that helps to encapsulate the entire piece or informs the reader about the topic, and a body that tells the entire story \cite{van1983discourse}. 

We rely on the MLSUM dataset \cite{scialom-etal-2020-mlsum}, which consists of news articles that have the structure mentioned above.
We use subhead and body sentences because the former has sentences summarizing the entire article, while the latter uses broader wording. Figure~\ref{fig.ssp} shows an example of each style.
We randomly sample one sentence from the subhead as a positive instance and one sentence from the body as a negative sample.
The task is a binary classification that takes the representation of the sentence as input.

\begin{figure}[!h]
\begin{center}
\fbox
{
\begin{minipage}{0.45\textwidth}
    \textit{\textbf{Subhead}}: Los Reyes presiden este sábado el desfile de las Fuerzas Armadas \newline
    \textit{\textbf{Body}}: Sevilla acoge este sábado el tradicional desfile de las Fuerzas Armadas, que estará presidido por los Reyes de España
\end{minipage}
}
\vspace{-0.2cm}
\caption{Examples of SSP. One sentence is from the subhead, while the other is from the body of a news.}
\vspace{-0.5cm}
\label{fig.ssp}
\end{center}
\end{figure}

\subsection{Discourse Relations (DR)}

A direct way to test discourse knowledge is to predict the relations between sentences, which is why the RST Discourse Treebank \cite{carlson-etal-2001-building} was used in previous work \cite{ferracane-etal-2019-evaluating,chen-etal-2019-evaluation}.
We consider the RST Spanish Treebank \cite{da-cunha-etal-2011-development} for our Spanish version, which consists of an annotated corpus with rhetorical relations.

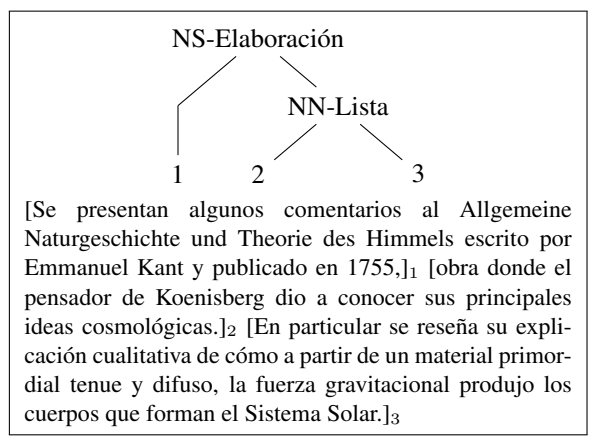
\begin{figure}[!h]
\begin{center}
\fbox
{
\begin{minipage}{0.45\textwidth}

\begin{center}
\begin{tikzpicture}[scale = 0.6, sibling distance=10em]
  \node {NS-Elaboración}
    child { child { node {1} } }
    child { node {NN-Lista} 
        child { node {2} }
        child { node {3} } };
\end{tikzpicture}
\end{center}
\vspace{-0.4cm}

\small
[Se presentan algunos comentarios al Allgemeine Naturgeschichte und Theorie des Himmels escrito por Emmanuel Kant y publicado en 1755,]$_1$ [obra donde el pensador de Koenisberg dio a conocer sus principales ideas cosmológicas.]$_2$ [En particular se reseña su explicación cualitativa de cómo a partir de un material primordial tenue y difuso, la fuerza gravitacional produjo los cuerpos que forman el Sistema Solar.]$_3$

\end{minipage}
}
\vspace{-0.2cm}
\caption{An RST Spanish Treebank tree with nucleus-satellite (NS) and multi-nuclear (NN) relations.}
\vspace{-0.5cm}
\label{fig.rst}
\end{center}
\end{figure}

According to RST \cite{MANN1988}, a text can be segmented into Elementary Discourse Units (EDUs) linked by means of nucleus-satellite (NS) or multi-nuclear (NN) rhetorical relations.
In the first, the satellite provides additional information about the nucleus, on which it depends (e.g., Fondo, Condición). In the second, several nuclei elements are connected at the same level, so no element is dependent on any other (e.g., Unión, Lista).
For instance, Figure~\ref{fig.rst} shows an example with a relation NS and NN. A relation can take multiple units, so like \newcite{chen-etal-2019-evaluation}, we rely on right-branching trees for non-binary relations to binarize the tree structure and use the 29 coarse-grained relations defined by \newcite{da-cunha-etal-2011-development}. We adopt the originally proposed training and testing splits.

\begin{table*}[]
\centering
\begin{tabular}{l|cccc|ccccc|}
\cline{2-10}
                                      & \multicolumn{4}{c|}{\textbf{SentEval}}                                                                                & \multicolumn{5}{c|}{\textbf{DiscoEval}}                                                                                                                    \\ \hline
\multicolumn{1}{|l|}{\textbf{Models}} & \multicolumn{1}{c|}{\textbf{SC}} & \multicolumn{1}{c|}{\textbf{SPC}} & \multicolumn{1}{c|}{\textbf{SS}} & \textbf{LPT} & \multicolumn{1}{c|}{\textbf{SP}} & \multicolumn{1}{c|}{\textbf{BSO}} & \multicolumn{1}{c|}{\textbf{DC}} & \multicolumn{1}{c|}{\textbf{SSP}} & \textbf{DR} \\ \hline
\multicolumn{1}{|l|}{Sent2Vec}        & \multicolumn{1}{c|}{\underline{75.11}}       & \multicolumn{1}{c|}{59.51}        & \multicolumn{1}{c|}{\textbf{76.05}}       & 66.89       & \multicolumn{1}{c|}{36.49}       & \multicolumn{1}{c|}{54.92}        & \multicolumn{1}{c|}{55.77}       & \multicolumn{1}{c|}{70.88}        & 36.69        \\
\multicolumn{1}{|l|}{ELMo}            & \multicolumn{1}{c|}{71.50}       & \multicolumn{1}{c|}{\textbf{61.62}}        & \multicolumn{1}{c|}{62.06}       & \underline{69.90}       & \multicolumn{1}{c|}{37.13}       & \multicolumn{1}{c|}{55.13}        & \multicolumn{1}{c|}{58.68}       & \multicolumn{1}{c|}{72.60}        & 45.14        \\ \hline
\multicolumn{1}{|l|}{ELECTRA}         & \multicolumn{1}{c|}{62.80}       & \multicolumn{1}{c|}{51.40}        & \multicolumn{1}{c|}{42.07}       & 64.20
   & \multicolumn{1}{c|}{38.56}       & \multicolumn{1}{c|}{56.85}        & \multicolumn{1}{c|}{55.18}       & \multicolumn{1}{c|}{76.22}        & 37.59        \\
\multicolumn{1}{|l|}{RoBERTa-BNE}     & \multicolumn{1}{c|}{72.51}       & \multicolumn{1}{c|}{54.57}        & \multicolumn{1}{c|}{41.34}       & 68.22       & \multicolumn{1}{c|}{\underline{41.82}}       & \multicolumn{1}{c|}{57.02}        & \multicolumn{1}{c|}{56.31}       & \multicolumn{1}{c|}{76.83}        & 39.21        \\
\multicolumn{1}{|l|}{BERTIN}          & \multicolumn{1}{c|}{73.54}       & \multicolumn{1}{c|}{55.47}        & \multicolumn{1}{c|}{32.53}       & 67.72       & \multicolumn{1}{c|}{41.66}       & \multicolumn{1}{c|}{56.66}        & \multicolumn{1}{c|}{55.54}       & \multicolumn{1}{c|}{\textbf{78.42}}        & 45.86        \\
\multicolumn{1}{|l|}{BETO}            & \multicolumn{1}{c|}{\textbf{76.34}}       & \multicolumn{1}{c|}{58.17}        & \multicolumn{1}{c|}{55.37}       & 69.38       & \multicolumn{1}{c|}{41.43}       & \multicolumn{1}{c|}{\underline{57.53}}        & \multicolumn{1}{c|}{\underline{60.89}}       & \multicolumn{1}{c|}{75.33}        & \underline{47.84}        \\
\multicolumn{1}{|l|}{mBERT}           & \multicolumn{1}{c|}{70.47}       & \multicolumn{1}{c|}{\underline{60.05}}        & \multicolumn{1}{c|}{\underline{67.77}}       & \textbf{71.41}       & \multicolumn{1}{c|}{\textbf{43.21}}       & \multicolumn{1}{c|}{\textbf{57.97}}        & \multicolumn{1}{c|}{\textbf{63.45}}       & \multicolumn{1}{c|}{\underline{77.80}}        & \textbf{51.08}        \\ \hline
\end{tabular}
\vspace{-0.2cm}
\caption{Results for Spanish SentEval and Spanish DiscoEval by group. The best performing model is in bold, and the runner up method is underlined. 
The reported numbers are accuracy, except SS, which is Pearson's correlation.
}
\vspace{-0.2cm}
\label{table:results}
\end{table*}

To evaluate the representations, we first encode all EDUs. We then use averaged EDU representations of subtrees as inputs. Formally, the input to the classifier is $[x_{\scriptsize\textit{left}}, x_{\scriptsize\textit{right}}, x_{\scriptsize\textit{left}} \odot x_{\scriptsize\textit{right}}, |x_{\scriptsize\textit{left}}- x_{\scriptsize\textit{right}}|]$ and the label is the relation of the node. $x_{\scriptsize\textit{left}}$ and $x_{\scriptsize\textit{right}}$ are vectors of the left and right subtrees respectively. For instance, the input for the label ``NS-Elaboración'' from Figure~\ref{fig.rst} is $x_{\scriptsize\textit{left}}=x_1$ and $x_{\scriptsize\textit{right}}=\frac{x_2+x_3}{2}$.

\section{Experiments}

\subsection{Setup}
\paragraph{Parameters} We adopt and adapt the original implementation of SentEval and DiscoEval for our Spanish version, so the same hyperparameters can be set.
We use the PyTorch version of the classifiers, Adam optimizer with a batch size of 64, and 4 training epochs for all of the experiments. 
\paragraph{Datasets} 
We tokenize each dataset with the spaCy tokenizer \cite{spacy2} and save all files using a common file format with UTF-8 encoding.

\subsection{Models}
We benchmark all of the main Spanish sentence encoders available to date to the best of our knowledge. Sent2Vec\footnote{\href{https://github.com/BotCenter/spanish-sent2vec}{https://github.com/BotCenter/spanish-sent2vec}} \cite{pagliardini-etal-2018-unsupervised} which is a bilinear model, ELMo \cite{che-etal-2018-towards} which is based on bidirectional RNNs. More recent and based on Transformer \cite{NIPS2017_3f5ee243}, we evaluate BETO \cite{canete2020spanish}, the Spanish version of BERT \cite{devlin-etal-2019-bert},  BERTIN\footnote{\href{https://huggingface.co/bertin-project}{https://huggingface.co/bertin-project}} and RoBERTa-BNE \cite{gut2021spanish}, two versions in Spanish of the RoBERTa model \cite{liu2020roberta}. Finally, ELECTRA\footnote{\href{https://chriskhanhtran.github.io/posts/electra-spanish/}{https://chriskhanhtran.github.io/posts/electra-spanish/}} \cite{Clark2020ELECTRA:} that was trained on a small piece of data as part of a tutorial.
We also include the multilingual BERT (mBERT) for further comparison.
We use the base version for all models except ELECTRA, which is a small version.

For evaluating Sent2Vec and ELMo, we use their final representation. 
For the Transformer-based models, as proposed by \newcite{chen-etal-2019-evaluation}, we use the average of each layer's special tokens \texttt{[CLS]} as the sentence representation.

\begin{figure*}[!ht]
\centering
  \begin{minipage}{0.45\textwidth}
    \includegraphics[width=8cm]{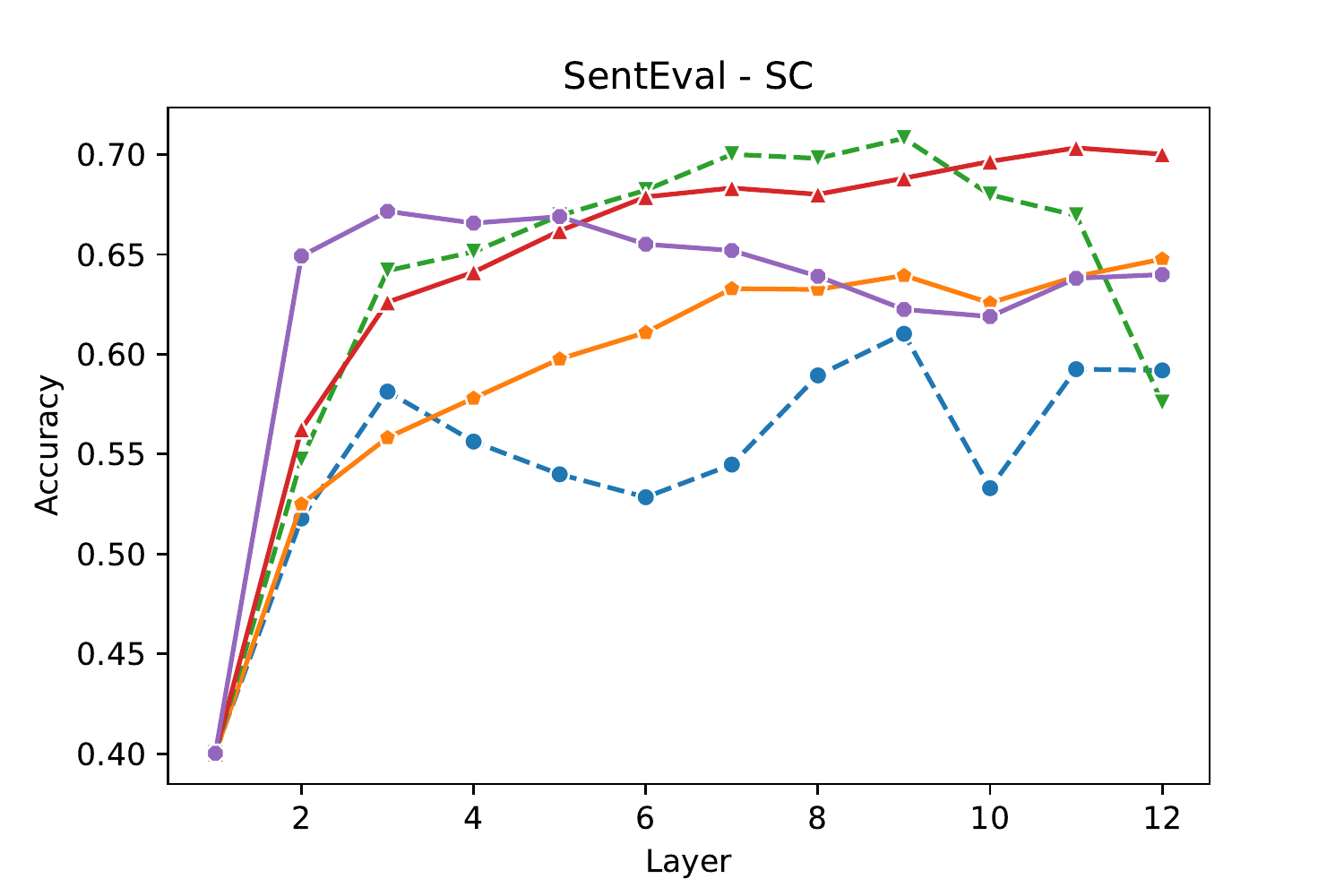}
  \end{minipage}
\hfill
  \begin{minipage}{0.45\textwidth}
    \includegraphics[width=8cm]{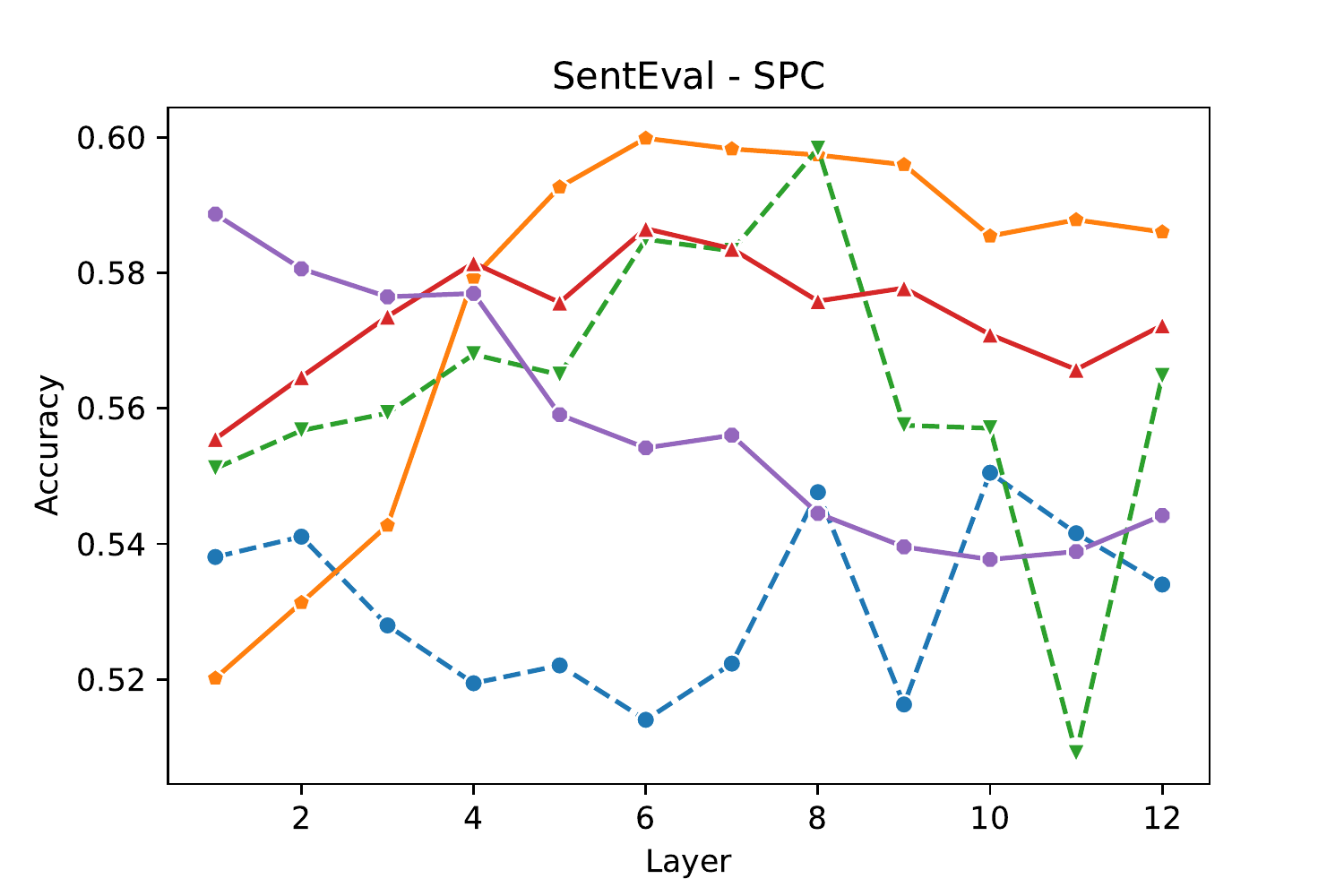}
  \end{minipage}
\hfill
  \begin{minipage}{0.45\textwidth}
    \includegraphics[width=8cm]{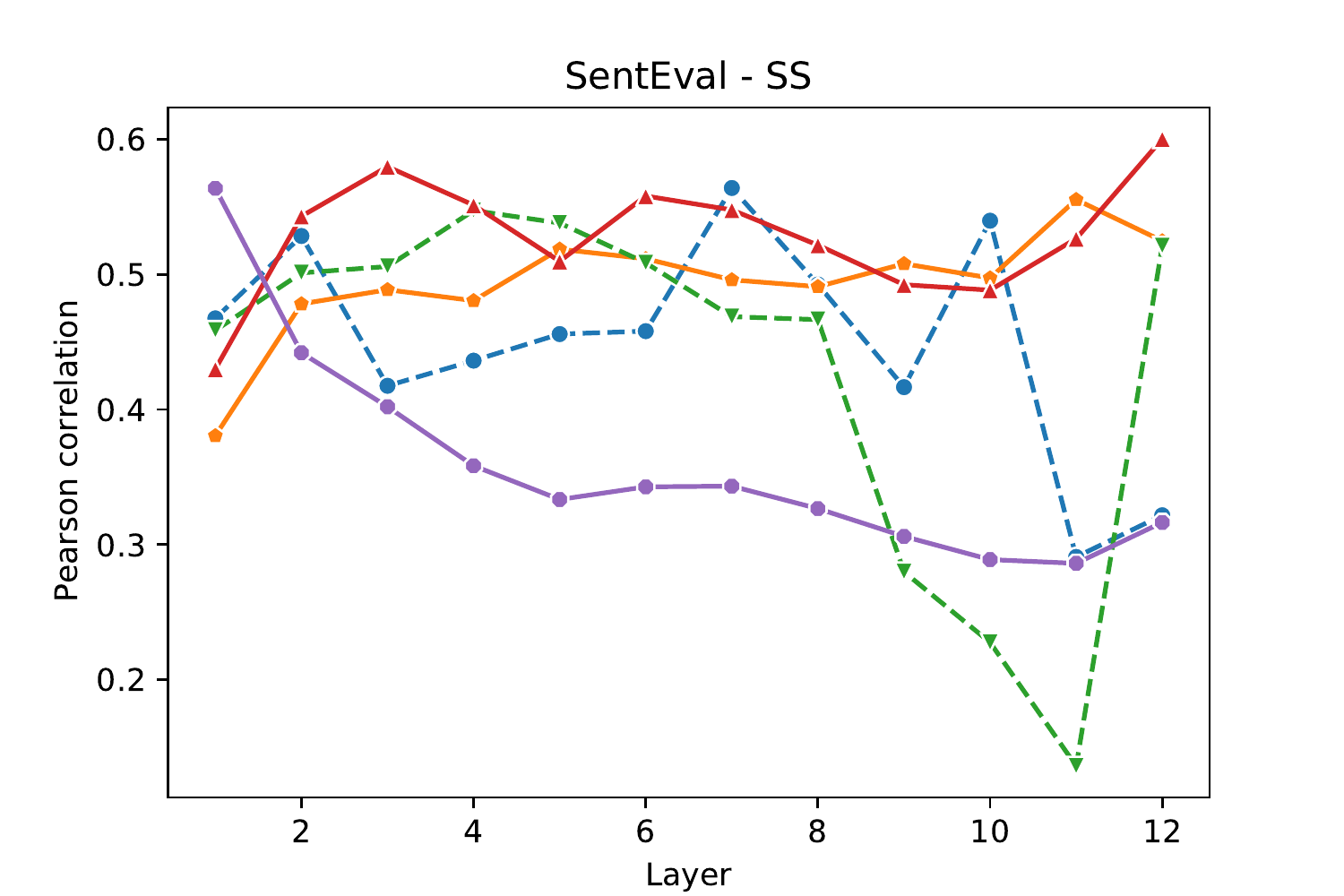}
  \end{minipage}
 \hfill
    \begin{minipage}{0.45\textwidth}
    \includegraphics[width=8cm]{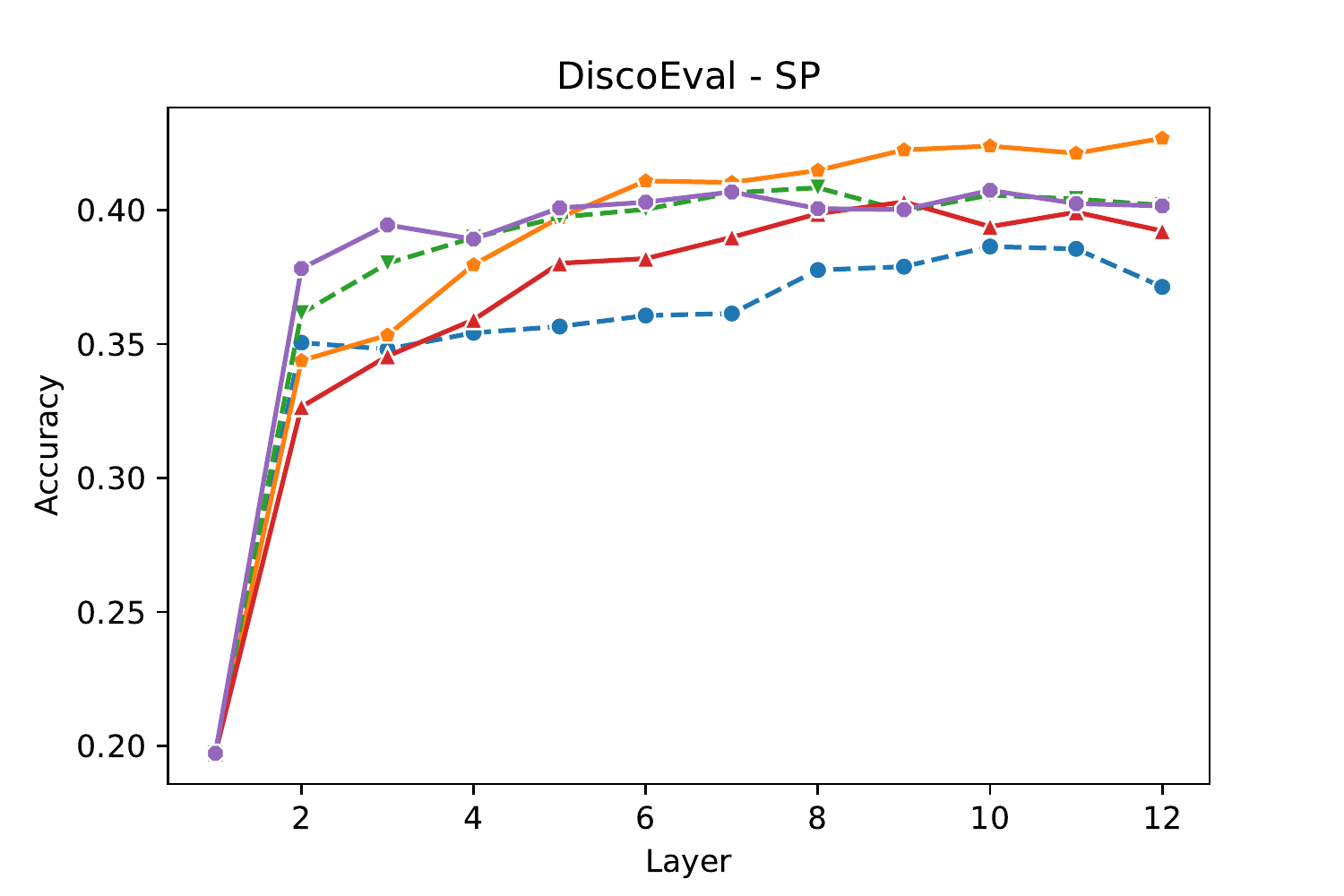}
  \end{minipage}
 \hfill 
    \begin{minipage}{0.45\textwidth}
    \includegraphics[width=8cm]{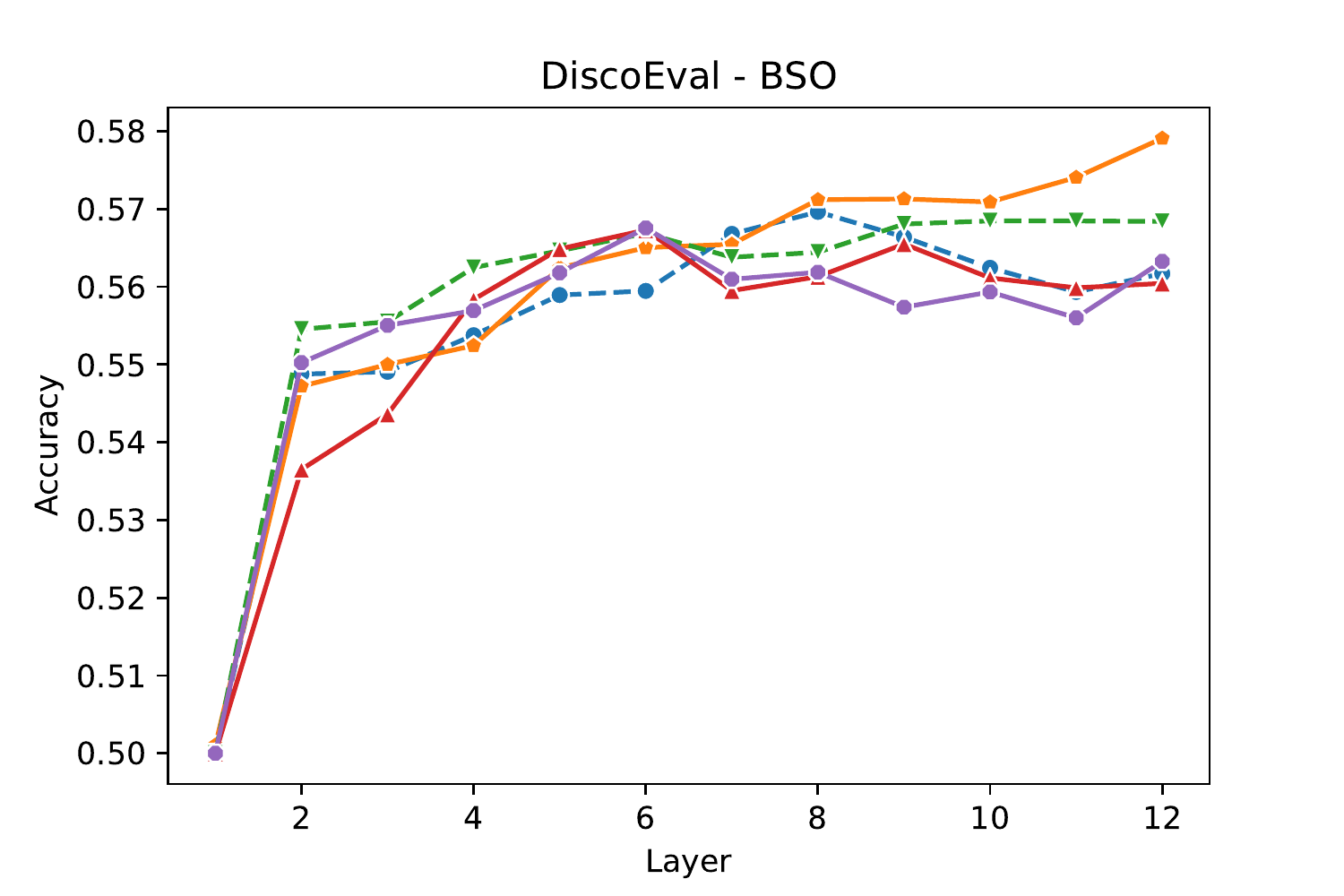}
  \end{minipage}
 \hfill 
    \begin{minipage}{0.45\textwidth}
    \includegraphics[width=8cm]{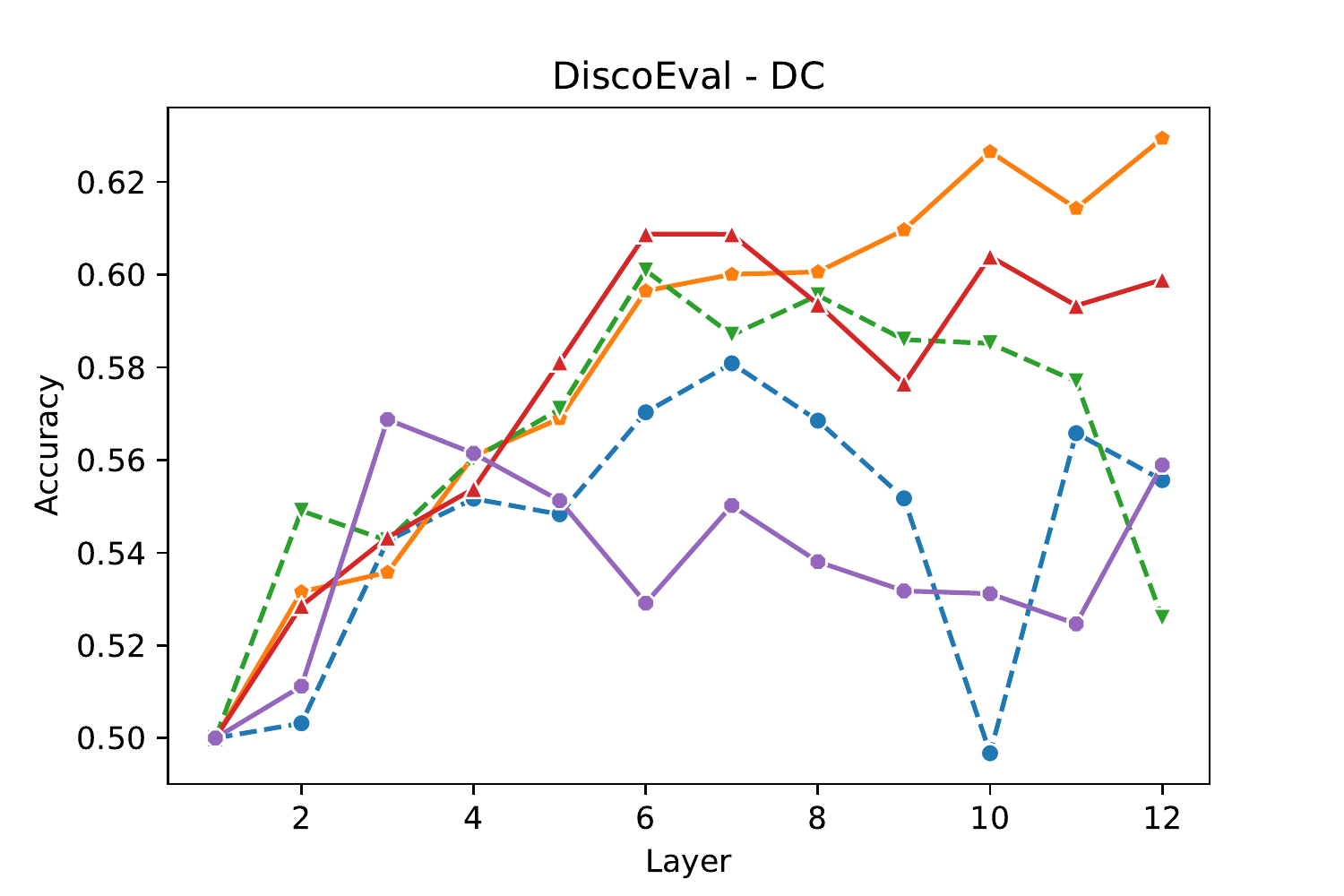}
  \end{minipage}
 \hfill 
    \begin{minipage}{0.45\textwidth}
    \includegraphics[width=8cm]{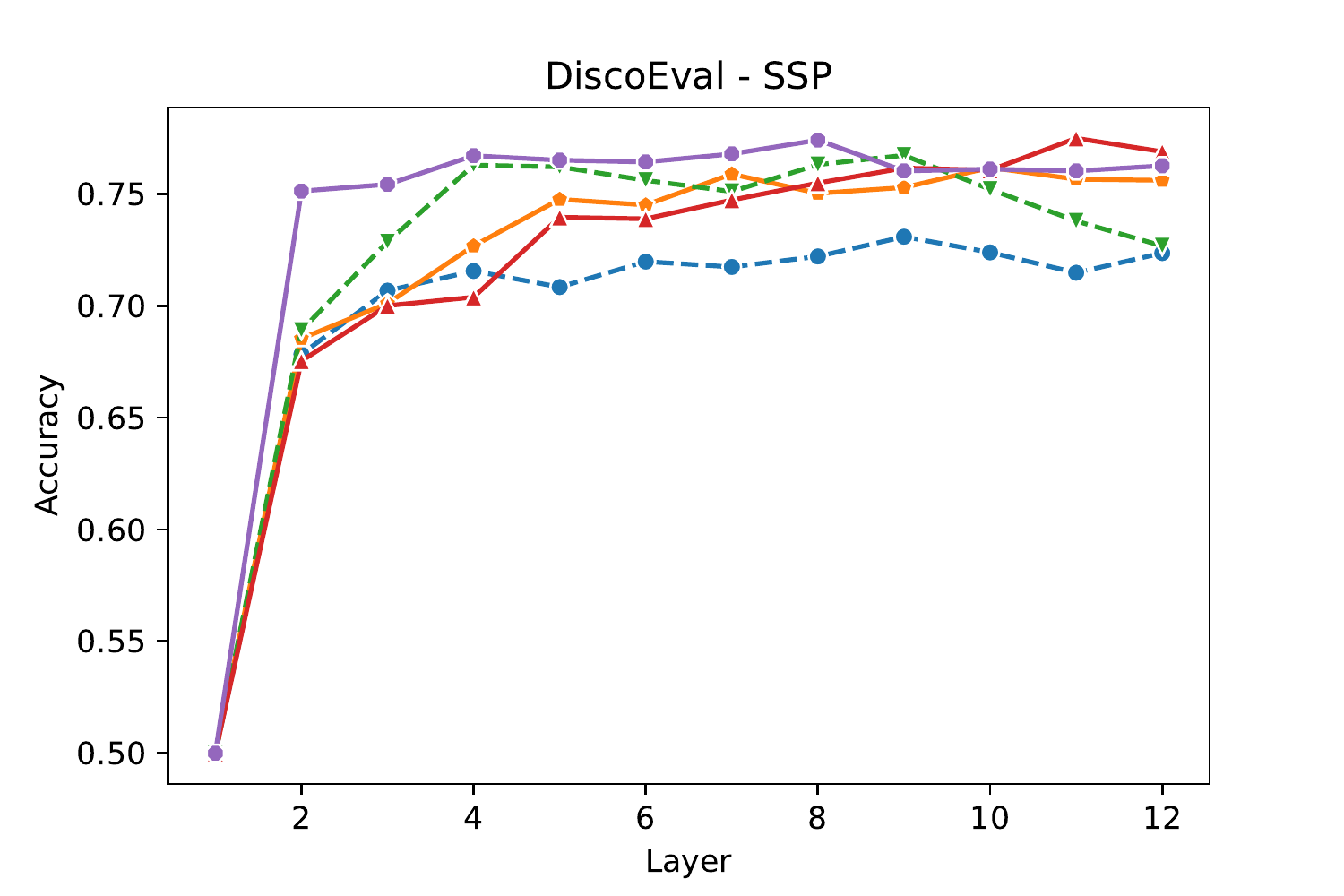}
  \end{minipage}
 \hfill 
    \begin{minipage}{0.45\textwidth}
    \includegraphics[width=8cm]{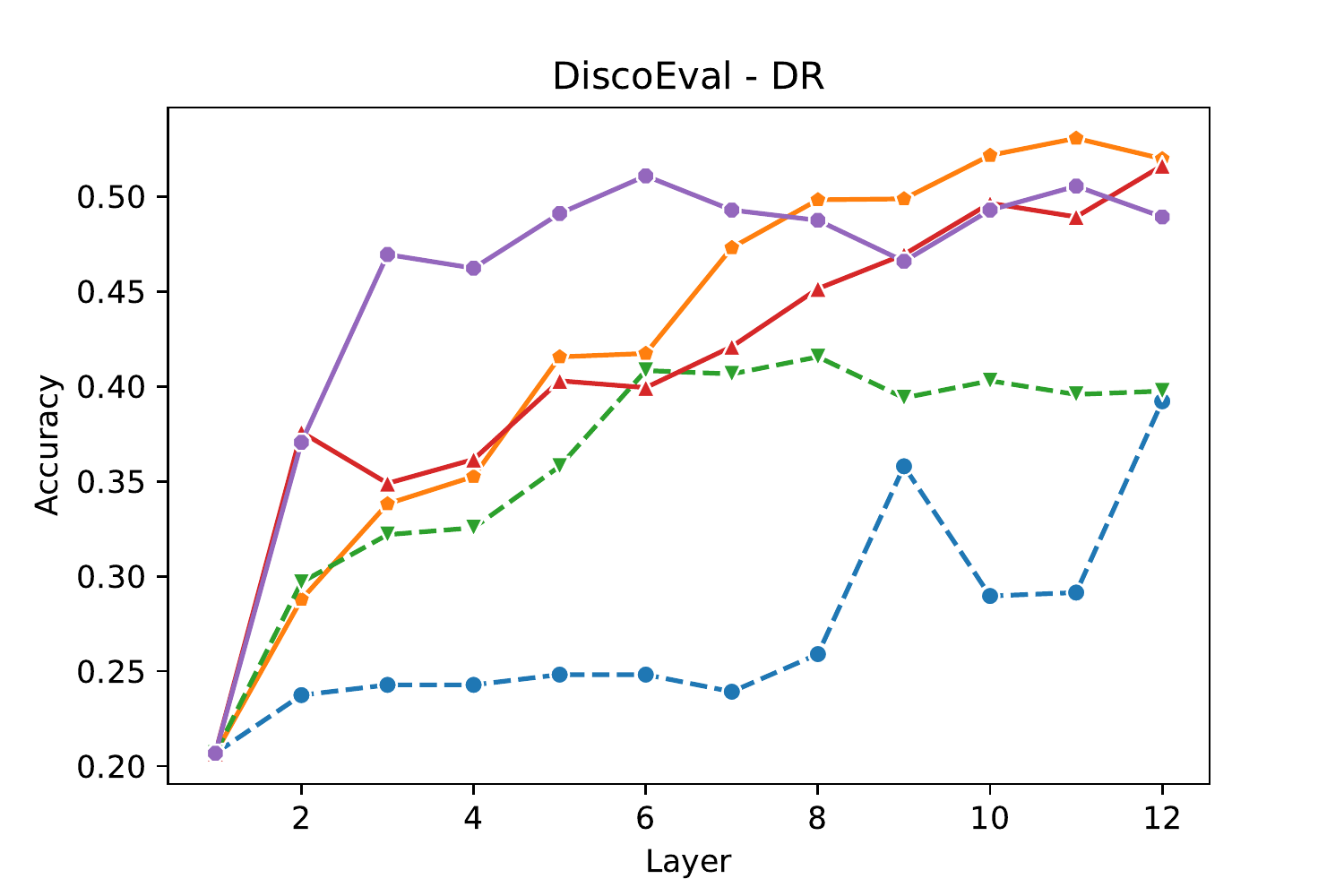}
  \end{minipage}
 \hfill 
    \begin{minipage}{0.60\textwidth}
    \centering
    \includegraphics[width=11cm]{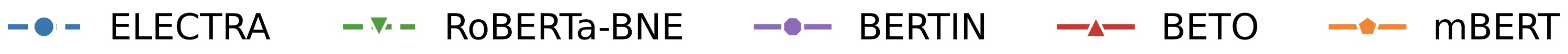}
  \end{minipage}
  
\vspace{-0.2cm}
\caption{Performance of DiscoEval and SentEval tasks using the \texttt{[CLS]} representation of layers 1 through 12.}
\vspace{-0.2cm}
\label{layerExperiments}
\end{figure*}

\subsection{Results}

Table \ref{table:results} shows the results of the experiments for all of the Spanish SentEval and Spanish DiscoEval tasks averaged for all of the datasets used for each of the tasks (for detailed results, see \hyperref[appendix:b]{Appendix B}).
It can be seen that, in general, the evaluation of all the language models' latent representations in both the Spanish SentEval and Spanish DiscoEval tasks show a similar behavior compared to their English language representations counterparts \cite{conneau-kiela-2018-senteval,chen-etal-2019-evaluation}. 
Regarding Spanish SentEval, for the SC tasks, the BETO model latent representation surpasses Sent2Vec, the second-best, by a 1.63 percentage difference (pd). This can be explained since BETO was trained with general domain sentences, capturing a representation capable of generalizing for any domain in the classification task. In general, SPC shows worse results than SC in terms of accuracy for all language models, where the best achieves an accuracy of 61.62\%. In the SPC task, it can be seen that ELMo learned representation surpasses the second-best representation (mBERT) by 2.6 pd. The SS task shows that Sent2Vec representation outperforms other representations by more than 12 pd in terms of Pearson's correlation, indicating that this learned representation can distinguish if a pair of sentences are semantically similar better than the representation learned by mBERT, which was not trained initially for this particular task. Finally, for the LPT task, the mBERT learned representation outperforms the BETO representation by 2.2 pd, showing that, when training a multi-language language model such as mBERT, the model can obtain richer sentence representations for a task that is more challenging than standard text classification.

Concerning the Spanish DiscoEval set of tasks, the SP task arises as one of the most challenging tasks, where mBERT, which is the best performing learned representation, reaches only a 43.21\% accuracy, improving by 3.33 pd over RoBERTa-BNE, the runner up model. We observe a similar pattern in the BSO task since the representation learned by mBERT outperforms the second-best model representation by a low margin of 0.7 pd, showing that training a Transformer-based model in multiple languages can obtain a richer representation for the task of ordering two sentences in a paragraph. A similar behavior is observed for the DC task, where the mBERT representation outperforms the runner-up method by more than 4 pd. The SSP task results indicate that the BERTIN learned representation surpasses mBERT by a small margin of 0.7 pd. Finally, for the RST task, mBERT representation shows the best performance in terms of accuracy compared to other language models, outperforming the BETO representation by 6.7 pd.
In general, it is observed that in most of the DiscoEval tasks, mBERT learned the best representation. The exception is for SSP, but the difference in accuracy compared to representation learned by BETO is small.
These results provide evidence that mBERT learns better representations when trained with multiple languages, allowing it to outperform other models on most probing tasks.

\subsection{Further Analysis}
In this section, we perform a per-layer performance analysis of the representations learned by transformer-based models. 
These experiments allow verifying which layers are more transferable for downstream tasks. Figure~\ref{layerExperiments} shows the results for all SentEval and DiscoEval groups.

It can be seen that the best performance fluctuates in the last layers, primarily between layers 10 and 12. Moreover, all representations perform well on the early layers for the SSP task, with accuracy levels near 0.7, indicating it is relatively straightforward. Nevertheless, all representations do not yield competitive performance for the SP task reaching a maximum accuracy slightly higher than 0.4, suggesting that they are ineffective at finding positions of a sentence in a discourse. Furthermore, something similar occurs with the DR task, where all representations achieve accuracy close to 0.5 for the last layers, showing that discovering relations between elements of discourse seems non-trivial to solve with the learned latent representations. 

Regarding the impact of the training data, we see that the representations generalize better when being trained with multiple languages than when using only Spanish text. Evidence of this is given by the performance of the representations learned by mBERT. In most cases, it outperforms other models' representations on several probing tasks. However, BETO representation beats the ones learned by mBERT in the last layers for SC, SS, and SSP, suggesting that for these tasks, representations learned only with Spanish texts seem to be more critical for obtaining an informative latent representation.

Another factor that positions mBERT and BETO as the two best-learned representations is that both were trained with more data, implying a better performance than ELECTRA and BERTIN, which were trained with fewer data. Interestingly, representations learned by RoBERTa-BNE do not get the desired performance compared to other representations, particularly on the early layers and on the last layers on tasks such as DC, DR, SSP, SC, and SPC. 

\section{Related Work}
\subsection{Language Model Evaluations}
We can find at least four approaches in the work carried out in the evaluation of language models.
The first focuses on evaluating the adaptability of a language model to a new domain through fine-tuning. GLUE \cite{wang-etal-2018-glue} and SuperGLUE \cite{NEURIPS2019_4496bf24} are examples of this approach that include several downstream tasks.
The second involves evaluating the generalization of text representations by incorporating a classifier for downstream tasks on top of them. Following this approach, SentEval \cite{conneau-kiela-2018-senteval} and DiscoEval \cite{chen-etal-2019-evaluation} include tasks at the sentence and discourse level.
The third focuses on stress tests \cite{naik-etal-2018-stress,aspillaga-etal-2020-stress,araujo-etal-2021-stress} that seek to assess the ability of language models to adapt to cases designed to confuse them.
The fourth objective is an evaluation from a linguistic perspective \cite{warstadt-etal-2019-investigating,10.1162/tacl_a_00298,puccetti-etal-2021-bert} to elucidate the models' actual linguistic capacities or knowledge.  

The aforementioned benchmarks are scarce for languages other than English. This, in fact, is the case for Spanish. For instance, regarding the adaptability evaluation for Spanish models, \newcite{canete2020spanish} recently proposed GLUES, a Spanish version of GLUE.
In the case of representation evaluation, most of the work is in a cross-linguistic setting for word \cite{10.1162/coli_a_00376}, sentence \cite{ravishankar-etal-2019-probing} and discourse \cite{koto-etal-2021-discourse} evaluations. For this reason and following the motivation of works such as RuSentEval \cite{mikhailov-etal-2021-rusenteval}, we provide SentEval and DiscoEval in Spanish, which consists of tasks originally created with texts in Spanish and aimed at evaluating models of that language.

\subsection{Sentence Encoders}
Pre-trained self-supervised language models have become the de facto sentence encoders. 
Early work in deep learning introduced ELMo \cite{peters-etal-2018-deep}.
With this model, sentence representations are produced by a mean-pooling of all contextualized word representations.
After the Transformer model \cite{NIPS2017_3f5ee243}, several models were proposed \cite{devlin-etal-2019-bert,liu2020roberta,Clark2020ELECTRA:}. These BERT-type models produce sentence representations using a special token \texttt{[CLS]}.
More recently, some models \cite{lee-etal-2020-slm,iter-etal-2020-pretraining,araujo-etal-2021-augmenting} have been proposed to improve discourse-level representations by incorporating additional components or mechanisms into the vanilla BERT.

Furthermore, due to the success of deep learning sentence encoders, some Spanish models were released.
\newcite{che-etal-2018-towards} released ELMo for many languages, including Spanish. BETO \cite{canete2020spanish} the Spanish version of BERT \cite{devlin-etal-2019-bert} was trained on a large Spanish corpus. RoBERTa-BNE \cite{gut2021spanish}, the Spanish version of the RoBERTa model \cite{liu2020roberta}, was trained on a corpus of crawled \textit{.es} domains.

\section{Conclusion}
We introduce Spanish SentEval and Spanish DiscoEval, two test suites for evaluating stand-alone and discourse-aware sentence representations. 
Like the English versions, our work aims to evaluate the representations of current and future Spanish language models.
Our benchmarks consist of a single pipeline that attempts a fair and less cumbersome assessment across multiple tasks with text from different domains.
As future work, more tasks could be included in these benchmarks. Likewise, other types of evaluations such as stress or linguistic tests could be carried out to evaluate the actual capacities of the language models taking into account the peculiarities of the Spanish language.

\section{Acknowledgements}
We want to thank the authors of the datasets for providing us access to their data.
This work was supported by the National Center for Artificial Intelligence CENIA FB210017, Basal ANID. Felipe Bravo-Marquez was supported by ANID FONDECYT grant 11200290, U-Inicia
VID Project UI-004/20 and ANID -Millennium Science Initiative Program - Code ICN17\_002.

\section{Bibliographical References}\label{reference}

\bibliographystyle{lrec2022-bib}
\bibliography{lrec2022-references}


\clearpage
\onecolumn
\section*{Appendix A: Details of Datasets}
\label{appendix:a}

\begin{table*}[!htbp]
\small
\centering
\setlength\tabcolsep{1.75mm}
\begin{tabular}{|l|c|l|c|llclc}
\cline{1-4} \cline{6-9}
\multicolumn{1}{|c|}{\textbf{Name}} & \textbf{N} & \multicolumn{1}{c|}{\textbf{Description}} & \textbf{C} & \multicolumn{1}{l|}{} & \multicolumn{1}{c|}{\textbf{Name}}  & \multicolumn{1}{c|}{\textbf{N}} & \multicolumn{1}{c|}{\textbf{Description}}        & \multicolumn{1}{c|}{\textbf{C}} \\ \cline{1-4} \cline{6-9} 
TASS1                               & 14.4k      & Polarity (2020 joint task 1)                & 3          & \multicolumn{1}{l|}{} & \multicolumn{1}{l|}{SP wiki/mlsum}  & \multicolumn{1}{c|}{18k}        & \multicolumn{1}{l|}{Sentence position}           & \multicolumn{1}{c|}{5}          \\ \cline{1-4} \cline{6-9} 
TASS2                               & 8.4k       & Emotion (2020 task 2)                 & 7          & \multicolumn{1}{l|}{} & \multicolumn{1}{l|}{SP thesis}      & \multicolumn{1}{c|}{14k}        & \multicolumn{1}{l|}{Sentence position}           & \multicolumn{1}{c|}{5}          \\ \cline{1-4} \cline{6-9} 
MC                                  & 2.6k       & Sentiment (Movie reviews)                 & 2          & \multicolumn{1}{l|}{} & \multicolumn{1}{l|}{BSO wiki/mlsum} & \multicolumn{1}{c|}{18k}        & \multicolumn{1}{l|}{Sentence ordering}           & \multicolumn{1}{c|}{2}          \\ \cline{1-4} \cline{6-9} 
FAC                                 & 5k         & Subjectivity/Objectivity                  & 2          & \multicolumn{1}{l|}{} & \multicolumn{1}{l|}{BSO thesis}     & \multicolumn{1}{c|}{14k}        & \multicolumn{1}{l|}{Sentence ordering}           & \multicolumn{1}{c|}{2}          \\ \cline{1-4} \cline{6-9} 
SQC                                 & 6.7k       & Question-type                             & 6          & \multicolumn{1}{l|}{} & \multicolumn{1}{l|}{DC wiki/opus}   & \multicolumn{1}{c|}{18k}        & \multicolumn{1}{l|}{Coherence}                   & \multicolumn{1}{c|}{2}          \\ \cline{1-4} \cline{6-9} 
PAWS-es                             & 51.4k      & Paraphase                                 & 2          & \multicolumn{1}{l|}{} & \multicolumn{1}{l|}{DC gdd}         & \multicolumn{1}{c|}{6.6k}       & \multicolumn{1}{l|}{Coherence}                   & \multicolumn{1}{c|}{2}          \\ \cline{1-4} \cline{6-9} 
NLI-es                              & 9.9k       & Entailment                                & 2          & \multicolumn{1}{l|}{} & \multicolumn{1}{l|}{SSP mlsum}      & \multicolumn{1}{c|}{18k}        & \multicolumn{1}{l|}{Section prediction}          & \multicolumn{1}{c|}{2}          \\ \cline{1-4} \cline{6-9} 
SICK-es                             & 9.8k       & Entailment/Relatedness                    & 2          & \multicolumn{1}{l|}{} & \multicolumn{1}{l|}{DR rst}         & \multicolumn{1}{c|}{3.3k}       & \multicolumn{1}{l|}{Rhetorical structure theory} & \multicolumn{1}{c|}{29}         \\ \cline{1-4} \cline{6-9} 
STS14                               & 804        & Semantic Similarity                       & score      &                       &                                     & \multicolumn{1}{l}{}            &                                                  &                                 \\ \cline{1-4}
STS15                               & 751        & Semantic Similarity                       & score      &                       &                                     & \multicolumn{1}{l}{}            &                                                  & \multicolumn{1}{l}{}            \\ \cline{1-4}
STS17                               & 250        & Semantic Similarity                       & score      &                       &                                     & \multicolumn{1}{l}{}            &                                                  & \multicolumn{1}{l}{}            \\ \cline{1-4}
\end{tabular}
\vspace{-0.2cm}
\caption{Details of the SentEval and DiscoEval datasets. N shows the number of instances and C the number of classes.}
\label{table:stas}
\vspace{-0.2cm}
\end{table*}


\section*{Appendix B: Full Results of Spanish Models}
\label{appendix:b}

\begin{table*}[!htbp]
\small
\centering
\setlength\tabcolsep{1.75mm}
\begin{tabular}{l|ccccc|ccc|cccc|}
\cline{2-13}
\textbf{}                             & \multicolumn{5}{c|}{\textbf{SC}}                                                                                                                                                                                                                                    & \multicolumn{3}{c|}{\textbf{SPC}}                                                                                                                                                                                                  & \multicolumn{4}{c|}{\textbf{SS}}                                                                                                                                                                                                                                                                                 \\ \hline
\multicolumn{1}{|l|}{\textbf{Models}} & \multicolumn{1}{c|}{\textbf{\begin{tabular}[c]{@{}c@{}}TASS\\ 1\end{tabular}}} & \multicolumn{1}{c|}{\textbf{\begin{tabular}[c]{@{}c@{}}TASS\\ 2\end{tabular}}} & \multicolumn{1}{l|}{\textbf{MC}} & \multicolumn{1}{c|}{\textbf{FAC}} & \textbf{SQC}               & \multicolumn{1}{c|}{\textbf{\begin{tabular}[c]{@{}c@{}}PAWS\\ -es\end{tabular}}} & \multicolumn{1}{c|}{\textbf{\begin{tabular}[c]{@{}c@{}}NLI\\ -es\end{tabular}}} & \textbf{\begin{tabular}[c]{@{}c@{}}SICK\\ -es-E\end{tabular}} & \multicolumn{1}{c|}{\textbf{\begin{tabular}[c]{@{}c@{}}SICK\\ -es-R\end{tabular}}} & \multicolumn{1}{c|}{\textbf{\begin{tabular}[c]{@{}c@{}}STS\\ 14\end{tabular}}} & \multicolumn{1}{c|}{\textbf{\begin{tabular}[c]{@{}c@{}}STS\\ 15\end{tabular}}} & \textbf{\begin{tabular}[c]{@{}c@{}}STS\\ 17\end{tabular}} \\ \hline
\multicolumn{1}{|l|}{Sent2Vec}        & \multicolumn{1}{l|}{60.91}                                                     & \multicolumn{1}{l|}{65.47}                                                     & \multicolumn{1}{l|}{75.94}       & \multicolumn{1}{l|}{95.46}        & \multicolumn{1}{l|}{77.78} & \multicolumn{1}{c|}{56.15}                                                       & \multicolumn{1}{c|}{48.35}                                                      & 74.03                                                         & \multicolumn{1}{c|}{78.98}                                                         & \multicolumn{1}{c|}{83.12}                                                     & \multicolumn{1}{c|}{60.02}                                                     & 82.10                                                     \\
\multicolumn{1}{|l|}{ELMo}            & \multicolumn{1}{l|}{56.07}                                                     & \multicolumn{1}{l|}{62.40}                                                     & \multicolumn{1}{l|}{66.91}       & \multicolumn{1}{l|}{94.66}        & \multicolumn{1}{l|}{77.48} & \multicolumn{1}{c|}{58.8}                                                        & \multicolumn{1}{c|}{49.28}                                                      & 76.78                                                         & \multicolumn{1}{c|}{75.56}                                                         & \multicolumn{1}{c|}{69.76}                                                     & \multicolumn{1}{c|}{45.88}                                                     & 57.02                                                     \\ \hline
\multicolumn{1}{|l|}{ELECTRA}         & \multicolumn{1}{l|}{58.04}                                                     & \multicolumn{1}{l|}{54.83}                                                     & \multicolumn{1}{l|}{65.61}       & \multicolumn{1}{l|}{94.18}        & 41.33                      & \multicolumn{1}{c|}{55.90}                                                       & \multicolumn{1}{c|}{36.39}                                                      & 61.90                                                         & \multicolumn{1}{c|}{56.40}                                                         & \multicolumn{1}{c|}{54.29}                                                     & \multicolumn{1}{c|}{30.85}                                                     & 26.74                                                     \\
\multicolumn{1}{|l|}{RoBERTa-BNE}     & \multicolumn{1}{l|}{64.10}                                                     & \multicolumn{1}{l|}{64.62}                                                     & \multicolumn{1}{l|}{75.11}       & \multicolumn{1}{l|}{96.66}        & 62.07                      & \multicolumn{1}{c|}{56.30}                                                       & \multicolumn{1}{c|}{40.60}                                                      & 66.82                                                         & \multicolumn{1}{c|}{67.79}                                                         & \multicolumn{1}{c|}{57.11}                                                     & \multicolumn{1}{c|}{34.13}                                                     & 6.31                                                      \\
\multicolumn{1}{|l|}{BERTIN}          & \multicolumn{1}{l|}{59.20}                                                     & \multicolumn{1}{l|}{61.86}                                                     & \multicolumn{1}{l|}{74.46}       & \multicolumn{1}{l|}{95.90}        & 76.30                      & \multicolumn{1}{c|}{56.75}                                                       & \multicolumn{1}{c|}{42.45}                                                      & 67.20                                                         & \multicolumn{1}{c|}{51.83}                                                         & \multicolumn{1}{c|}{31.24}                                                     & \multicolumn{1}{c|}{18.67}                                                     & 28.38                                                     \\
\multicolumn{1}{|l|}{BETO}            & \multicolumn{1}{l|}{62.95}                                                     & \multicolumn{1}{l|}{64.44}                                                     & \multicolumn{1}{l|}{77.01}       & \multicolumn{1}{l|}{97.60}        & 79.70                      & \multicolumn{1}{c|}{57.65}                                                       & \multicolumn{1}{c|}{46.39}                                                      & 70.46                                                         & \multicolumn{1}{c|}{63.24}                                                         & \multicolumn{1}{c|}{65.84}                                                     & \multicolumn{1}{c|}{47.36}                                                     & 45.01                                                     \\
\multicolumn{1}{|l|}{mBERT}           & \multicolumn{1}{l|}{56.28}                                                     & \multicolumn{1}{l|}{58.02}                                                     & \multicolumn{1}{l|}{70.00}       & \multicolumn{1}{l|}{96.92}        & \multicolumn{1}{l|}{71.11} & \multicolumn{1}{c|}{57.50}                                                       & \multicolumn{1}{c|}{48.63}                                                      & 74.01                                                         & \multicolumn{1}{c|}{67.77}                                                         & \multicolumn{1}{c|}{62.14}                                                     & \multicolumn{1}{c|}{43.96}                                                     & 36.71                                                     \\ \hline
\end{tabular}
\vspace{-0.2cm}
\caption{Results of the SentEval tasks for each dataset (SC, SPC, SS).}
\label{table:sentresults1}
\vspace{-0.2cm}
\end{table*}

\begin{table*}[!htbp]
\small
\centering
\setlength\tabcolsep{1.75mm}
\begin{tabular}{l|ccccccccc|}
\cline{2-10}
\textbf{}                             & \multicolumn{9}{c|}{\textbf{LPT}}                                                                                                                                                                                                                                                                                                                                                                                                                                                                                                                                                                                                          \\ \hline
\multicolumn{1}{|l|}{\textbf{Models}} & \multicolumn{1}{c|}{\textbf{SentLen}} & \multicolumn{1}{c|}{\textbf{WC}} & \multicolumn{1}{c|}{\textbf{\begin{tabular}[c]{@{}c@{}}Tree\\ Depth\end{tabular}}} & \multicolumn{1}{c|}{\textbf{BiShift}} & \multicolumn{1}{c|}{\textbf{Tense}} & \multicolumn{1}{c|}{\textbf{SubjNum}} & \multicolumn{1}{c|}{\textbf{ObjNum}} & \multicolumn{1}{c|}{\textbf{SOMO}} & \textbf{CoordInv} \\ \hline
\multicolumn{1}{|l|}{Sent2Vec}        & \multicolumn{1}{c|}{73.99}           & \multicolumn{1}{c|}{4.98}                                                            & \multicolumn{1}{c|}{41.25}          & \multicolumn{1}{c|}{81.76}                                                           & \multicolumn{1}{c|}{87.72}          & \multicolumn{1}{c|}{83.49}                                                          & \multicolumn{1}{c|}{78.50}                                                         & \multicolumn{1}{c|}{51.64}                                                         & 74.44                                                                     \\
\multicolumn{1}{|l|}{ELMo}            & \multicolumn{1}{c|}{44.22}           & \multicolumn{1}{c|}{66.43}                                                           & \multicolumn{1}{c|}{49.84}          & \multicolumn{1}{c|}{78.47}                                                           & \multicolumn{1}{c|}{94.65}          & \multicolumn{1}{c|}{87.27}                                                          & \multicolumn{1}{c|}{80.21}                                                         & \multicolumn{1}{c|}{53.76}                                                         & 74.25                                                                     \\ \hline
\multicolumn{1}{|l|}{ELECTRA}         & \multicolumn{1}{c|}{73.99}           & \multicolumn{1}{c|}{4.98}                                                            & \multicolumn{1}{c|}{41.25}          & \multicolumn{1}{c|}{81.76}                                                           & \multicolumn{1}{c|}{87.72}          & \multicolumn{1}{c|}{83.49}                                                          & \multicolumn{1}{c|}{78.50}                                                         & \multicolumn{1}{c|}{51.64}                                                         & 74.44                                                                     \\
\multicolumn{1}{|l|}{RoBERTa-BNE}     & \multicolumn{1}{c|}{77.27}           & \multicolumn{1}{c|}{22.60}                                                           & \multicolumn{1}{c|}{44.80}          & \multicolumn{1}{c|}{77.07}                                                           & \multicolumn{1}{c|}{93.63}          & \multicolumn{1}{c|}{87.15}                                                          & \multicolumn{1}{c|}{79.65}                                                         & \multicolumn{1}{c|}{51.98}                                                         & 79.83                                                                     \\
\multicolumn{1}{|l|}{BERTIN}          & \multicolumn{1}{c|}{75.33}           & \multicolumn{1}{c|}{20.84}                                                           & \multicolumn{1}{c|}{45.02}          & \multicolumn{1}{c|}{73.40}                                                           & \multicolumn{1}{c|}{93.36}          & \multicolumn{1}{c|}{89.71}                                                          & \multicolumn{1}{c|}{73.92}                                                         & \multicolumn{1}{c|}{53.80}                                                         & 84.06                                                                     \\
\multicolumn{1}{|l|}{BETO}            & \multicolumn{1}{c|}{68.85}           & \multicolumn{1}{c|}{41.60}                                                           & \multicolumn{1}{c|}{45.42}          & \multicolumn{1}{c|}{75.72}                                                           & \multicolumn{1}{c|}{93.26}          & \multicolumn{1}{c|}{87.30}                                                          & \multicolumn{1}{c|}{77.99}                                                         & \multicolumn{1}{c|}{51.03}                                                         & 83.27                                                                     \\
\multicolumn{1}{|l|}{mBERT}           & \multicolumn{1}{c|}{75.41}           & \multicolumn{1}{c|}{41.27}                                                           & \multicolumn{1}{c|}{46.50}          & \multicolumn{1}{c|}{76.34}                                                           & \multicolumn{1}{c|}{95.30}          & \multicolumn{1}{c|}{89.57}                                                          & \multicolumn{1}{c|}{78.22}                                                         & \multicolumn{1}{c|}{54.10}                                                         & 86.02                                                                     \\ \hline
\end{tabular}
\vspace{-0.2cm}
\caption{Results of the SentEval tasks for each dataset (LPT).}
\label{table:sentresults2}
\vspace{-0.2cm}
\end{table*}

\begin{table*}[!htbp]
\small
\centering
\setlength\tabcolsep{1.75mm}
\begin{tabular}{l|ccc|ccc|ccc|c|c|}
\cline{2-12}
\textbf{}       & \multicolumn{3}{c|}{\textbf{SP}}                                                           & \multicolumn{3}{c|}{\textbf{BSO}}                                                          & \multicolumn{3}{c|}{\textbf{DC}}                                                       & \textbf{SSP}   & \textbf{DR} \\ \hline
\multicolumn{1}{|l|}{\textbf{Models}} & \multicolumn{1}{c|}{\textbf{wiki}} & \multicolumn{1}{c|}{\textbf{mlsum}} & \textbf{thesis} & \multicolumn{1}{c|}{\textbf{wiki}} & \multicolumn{1}{c|}{\textbf{mlsum}} & \textbf{thesis} & \multicolumn{1}{c|}{\textbf{wiki}} & \multicolumn{1}{c|}{\textbf{opus}} & \textbf{gdd} & \textbf{mlsum} & \textbf{rst}    \\ \hline
\multicolumn{1}{|l|}{Sent2Vec}        & \multicolumn{1}{c|}{42.77}         & \multicolumn{1}{c|}{30.40}          & 36.29           & \multicolumn{1}{c|}{51.56}         & \multicolumn{1}{c|}{50.46}          & 62.75           & \multicolumn{1}{c|}{66.67}         & \multicolumn{1}{c|}{49.85}         & 50.78        & 70.88          & 36.69        \\
\multicolumn{1}{|l|}{ELMo}            & \multicolumn{1}{c|}{44.50}         & \multicolumn{1}{c|}{29.88}          & 37.01           & \multicolumn{1}{c|}{51.42}         & \multicolumn{1}{c|}{50.89}          & 63.07           & \multicolumn{1}{c|}{68.73}         & \multicolumn{1}{c|}{52.62}         & 54.69        & 72.60          & 45.14        \\ \hline
\multicolumn{1}{|l|}{ELECTRA}         & \multicolumn{1}{c|}{45.65}         & \multicolumn{1}{c|}{31.45}          & 38.57           & \multicolumn{1}{c|}{52.44}         & \multicolumn{1}{c|}{52.83}          & 65.29           & \multicolumn{1}{c|}{57.88}         & \multicolumn{1}{c|}{50.62}         & 57.03        & 76.22          & 37.59        \\
\multicolumn{1}{|l|}{RoBERTa-BNE}     & \multicolumn{1}{c|}{47.58}         & \multicolumn{1}{c|}{35.10}          & 42.78           & \multicolumn{1}{c|}{52.99}         & \multicolumn{1}{c|}{51.04}          & 67.04           & \multicolumn{1}{c|}{68.12}         & \multicolumn{1}{c|}{50.80}         & 50.00        & 76.83          & 39.21        \\
\multicolumn{1}{|l|}{BERTIN}          & \multicolumn{1}{c|}{47.90}         & \multicolumn{1}{c|}{33.38}          & 43.71           & \multicolumn{1}{c|}{52.09}         & \multicolumn{1}{c|}{49.90}          & 67.99           & \multicolumn{1}{c|}{60.75}         & \multicolumn{1}{c|}{53.52}         & 52.34        & 78.42          & 45.86        \\
\multicolumn{1}{|l|}{BETO}            & \multicolumn{1}{c|}{48.90}         & \multicolumn{1}{c|}{32.35}          & 43.03           & \multicolumn{1}{c|}{53.50}         & \multicolumn{1}{c|}{51.94}          & 67.15           & \multicolumn{1}{c|}{75.62}         & \multicolumn{1}{c|}{53.15}         & 53.91        & 75.33          & 47.84        \\
\multicolumn{1}{|l|}{mBERT}           & \multicolumn{1}{c|}{51.95}         & \multicolumn{1}{c|}{33.23}          & 44.46           & \multicolumn{1}{c|}{53.79}         & \multicolumn{1}{c|}{51.75}          & 68.36           & \multicolumn{1}{c|}{80.00}         & \multicolumn{1}{c|}{53.33}         & 57.03        & 77.80          & 51.08        \\ \hline
\end{tabular}
\vspace{-0.2cm}
\caption{Results of the DiscoEval tasks for each dataset.}
\label{table:discoresults}
\vspace{-0.2cm}
\end{table*}

\end{document}